# A Formal Ontology-Based Classification of Lexemes and its Applications *

**Sreekavitha Parupalli and Navjyoti Singh**
Center for Exact Humanities (CEH)
International Institute of Information Technology, Hyderabad
sreekavitha.parupalli@research.iiit.ac.in
navjyoti@iiit.ac.in

## Abstract

The paper describes the enrichment of OntoSenseNet - a verb-centric lexical resource for Indian Languages. A major contribution of this work is preservation of an authentic Telugu dictionary by developing a computational version of the same. It is important because native speakers can better annotate the sense-types when both the word and its meaning are in Telugu. Hence efforts are made to develop the aforementioned Telugu dictionary and annotations are done manually. The manually annotated gold standard corpus consists 8483 verbs, 253 adverbs and 1673 adjectives. Annotations are done by native speakers according to defined annotation guidelines. In this paper, we provide an overview of the annotation procedure and present the validation of the developed resource through inter-annotator agreement. Additional words from Telugu WordNet are added to our resource and are crowd-sourced for annotation. The statistics are compared with the sense-annotated lexicon, our resource for more insights.

## 1 Introduction

Lexically rich resources form the foundation of all natural language processing(NLP) tasks. Maintaining the quality of resources is thus a high priority issue (Chatterjee et al., 2010). Hence, it is important to enhance and maintain the lexical resources of any language. This is of significantly more importance in case of resource poor languages like Telugu (Sravanthi et al., 2015).

Telugu WordNet is developed as a part of IndoWordNet at CFILT (Bhattacharyya, 2010), which is considered as the most exhaustive set of multilingual lexical assets for Indian languages. It consists of 21091 synsets in total. Telugu WordNet captures several other semantic relations such as hypernymy, hyponymy, holonymy, meronymy, antonymy.

## 2 Approach

Vagueness in word sense emerges when a specific word has multiple conceivable senses. Finding the right sense requires exhaustive information of words. The meaning of a word, from ontological viewpoint, can be understood based on its participation in classes, events and relations. A formal ontology is developed to computationally manipulate language at the level of meanings which have an intrinsic form (Otra, 2015). There is 7 sense-type classification of verbs and 4 sense-class classification of adverbs. Adjectives are identified as 12 sense-types. However these are reduced to 6 pairs. Further classification of adjectives, spatio-temporal classification, is developed.

## 3 Data Collection

Telugu is a Dravidian language native to India. It stands alongside Hindi, English and Bengali as one of the few languages with official primary language status in India[1]. Telugu language ranks third in the population with number of native speakers in India (74 million, 2001 census)[2]. However, the amount of lexical annotated resources available is considerably low. This deters the novelty of research possible in the language. Additionally, the properties of Telugu are significantly different compared to major languages such as English.

---
[1] https://en.wikipedia.org/wiki/Telugu_language
[2] https://web.archive.org/web/20131029190612/http://censusindia.gov.in/Census_Data_2001/Census_Data_Online/Language/

| Resource | Verb | Adverb | Adj |
|---|---|---|---|
| OntoSenseNet | 8483 | 253 | 1673 |
| Telugu WordNet[4] | 2803 | 477 | 5827 |
| Synsets in WordNet | 2795 | 442 | 5776 |
| Telugu-Hindi[5] | 9939 | 142 | 1253 |
| English-Telugu[6] | 4657 | 1893 | 6695 |

Table 1: Statistics of available lexical resources

| Sense-Type | OntoSenseNet | WordNet |
|---|---|---|
| Know | 7.8% | 6.5% |
| Do | 33.7% | 44.2% |
| Move | 26.7% | 14.3% |
| Be | 9.8% | 10.2% |
| Have | 13.7% | 15.1% |
| Cut | 3.9% | 5.1% |
| Cover | 4.4% | 4.6% |

Table 2: Sense-Type Classification of Verbs

Furthermore, till date there is no generally accessible dictionary reference till date. In this work, a Telugu lexicon is created manually from an existing old (first volume was printed in 1936), authentic dictionary 'శ్రీ సూర్యరాయాం-ధ్ర తెలుగు నిఘంటువు (Srī sūryarāyāṁdhra Telugu nighaṃṭuvu)' which has 8 volumes in total (Pantulu, 1988). Nearly 21,000 root words alongside their meanings were recorded. The resource is developed to enrich OntoSenseNet[3] with addition of regional language resources. For each word extracted, based on its meaning, sense-type was identified by native speakers of language. We are presenting some statistics of available resources in Table 1. There are around 36,000 words in the dictionary we developed whereas IndoWordNet lists 21,091 words. Even without further analysis and classification we can see that this resource enriches WordNet by adding almost 15,000 words. This was the motivation to start this work.

### 3.1 Validation of the Resource

Cohen's Kappa (Cohen, 1968) was used to measure inter coder agreement which proves the reliability. The annotations are done by one human expert and it is cross-checked by another annotator who is equally proficient. Both the annotators are native speakers of the language. Verbs and adverbs are randomly selected from our resource for the evaluation sample. The inter coder agreement for 500 Telugu verbs is 0.86 and for 100 Telugu adverbs it is 0.94. Validation of the language resource shows high agreement (Landis and Koch, 1977). However, further validation of the resource is still in progress.

## 4 Annotation Procedure

Each verb can have all the seven meaning primitives (sense-types) in it, in various degrees. The degree depends on the usage/popularity of a meaning in a language that leads to a particular sense-type annotation. In our resource we have identified two sense-types of each verb, i.e. primary and secondary. Entire lexicon of verbs and adverbs was classified however work is in progress for adjectives. Till date, 1673 adjectives were annotated. All of the annotations were done manually by native speakers of language in accordance with the classification presented in Section 2.

### 4.1 Crowd sourcing

Similar annotation was done on the synsets of WordNet. Out of 2795 verb synsets, we extracted a bag of words which are not present in the resource in use. We annotated each lexeme of these synsets as an separate entry. This set of lexemes are crowd-sourced and annotations are done following the annotation guidelines by six language experts. These words were divided into three sets and each set was annotated by two language experts separately. The aggregate value Cohen's Kappa was measured to be 0.91. We can observe that all the lexemes in synsets don't share the same primary sense-type. Another observation is that having sets that share the same primary and secondary sense-types would result in better WSD for tasks like machine translation.

Further observations are presented in Table 2. The table depicts the difference in Sense-

---

Statement1.htm

[3] http://ceh.iiit.ac.in/lexical_resource/index.html

Type distribution of verbs in OntoSenseNet for Telugu, IndoWordNet. This gives an overview (percentage distribution) of which sense-type is predominant in the above discussed lexical resources.

## 4.2 Example

ID :: 3434
CAT :: verb
CONCEPT :: ప్రతిరోజు సూర్యుడు తూర్పున రావడం (pratiroju sūryuḍu tūrpuna rāvaḍaṃ)
EXAMPLE :: సూర్యుడు తూర్పున ఉదయిస్తాడు (sūryuḍu tūrpuna udayistāḍu)
SYNSET-TELUGU :: ఉదయించు, పుట్టు, పొడతెంచు, అవతరించు, ఆవిర్భవించు, ఉద్భవించు, జనించు, జనియించు, ప్రభవించు, వచ్చు, ఏతెంచు.

In synset ID 3434, the verb puṭṭu (birth) is used in the sense of Sun rising in the east. In a sense that sun is taking birth i.e it conveys that sun came into existence. The primary sense of this would be Before|After as it deals with transition. Secondary sense would be Locus|Located as it shows the state of a sun in the dawn.

However, another (synonymous) word, janiṃcu (birth), in the synset is used to describe the birth of a child. In a sense that a mother gave birth to her child. This process of child-birth needs an agent hence the primary sense becomes Means|End as the action needs agent for it's accomplishment. The secondary sense would be Part|Whole as the child was separated from a whole i.e. his mother.

In this case, words from same synset have different primary sense-type. There was a high potential for such occurrences hence each word in the synset was considered as a new entry for the annotation task rather than assigning same primary and secondary senses to all the words in a synset.

## 5 Conclusion

In this paper, a manually sense-annotated lexicon is developed. Classification is done by expert native Telugu speakers. The validation of this resource is done using Cohen's Kappa that shows higher agreement. Further validation and enrichment of the resource is in progress. Classification of words in WordNet, that are not in the resource, is also attempted. A potential application of this resource has been discussed briefly.

## 6 Acknowledgements

This work is part of the ongoing MS thesis in Exact Humanities under the guidance of Prof. Radhika Mamidi, Prof. Navjyoti Singh. I am immensely grateful to Vijaya Lakshmi for helping me with data collection and annotation process of the whole resource. I would also like to show my gratitude to K Jithendra Babu, Historian & Chairman at Deccan Archaeological and Cultural Research Institute, Hyderabad for providing the hard copy of the dictionary (all the 8 volumes most of which are unavailable currently) and for sharing his pearls of wisdom with us during the course of this research. I thank my fellow researchers from LTRC lab, IIIT-H who provided insight and expertise that greatly assisted the research.